\documentclass{article}
 
\usepackage{microtype}
\usepackage{graphicx}
\usepackage{subfigure}
\usepackage{booktabs}

\usepackage{amsmath}
\usepackage{amsfonts}
\usepackage{graphicx}
\graphicspath{ {./images/} }
\usepackage{xcolor}

\usepackage{hyperref}

\usepackage[accepted]{icml2020}

\begin{document}

\twocolumn[
\icmltitle{Structured Weight Priors for Convolutional Neural Networks}

\icmlsetsymbol{equal}{*}

\begin{icmlauthorlist}
\icmlauthor{Tim Pearce}{cam}
\icmlauthor{Andrew Y.~K.~Foong}{cam}
\icmlauthor{Alexandra Brintrup}{cam}
\end{icmlauthorlist}

\icmlaffiliation{cam}{University of Cambridge}

\icmlcorrespondingauthor{Tim Pearce}{tp424@cam.ac.uk}

\icmlkeywords{Machine Learning, ICML, structure, weight priors, priors, architecture, Bayesian, neural networks, deep learning}

\vskip 0.3in
]

\printAffiliationsAndNotice{}

\begin{abstract}
Selection of an architectural prior well suited to a task (e.g. convolutions for image data) is crucial to the success of deep neural networks (NNs). Conversely, the weight priors within these architectures are typically left vague, e.g.~independent Gaussian distributions, which has led to debate over the utility of Bayesian deep learning. This paper explores the benefits of adding structure to weight priors. It initially considers first-layer filters of a convolutional NN, designing a prior based on random Gabor filters. Second, it considers adding structure to the prior of final-layer weights by estimating how each hidden feature relates to each class. Empirical results suggest that these structured weight priors lead to more meaningful functional priors for image data. This contributes to the ongoing discussion on the importance of weight priors.
\end{abstract}


\section{Introduction}

Convolutional neural networks (CNNs) can be viewed as constrained fully-connected NNs (fc NNs), where a particular pattern of weight pruning and sharing has been enforced. Despite CNNs being less expressive than fc NNs, they perform better when learning from finite training data on spatial tasks. This is because they appropriately trade off model flexibility (bias) with the ease of finding a set of weights that generalises (variance) \citep{LeCun1989}.
This is an example of an `\textbf{architectural prior}' - assumptions about the input to output mapping have been baked into the computational graph of the NN (e.g. translational invariance to features).

\begin{figure}[b]
\vskip -0.15in
\begin{center}
\centerline{\includegraphics[width=\columnwidth]{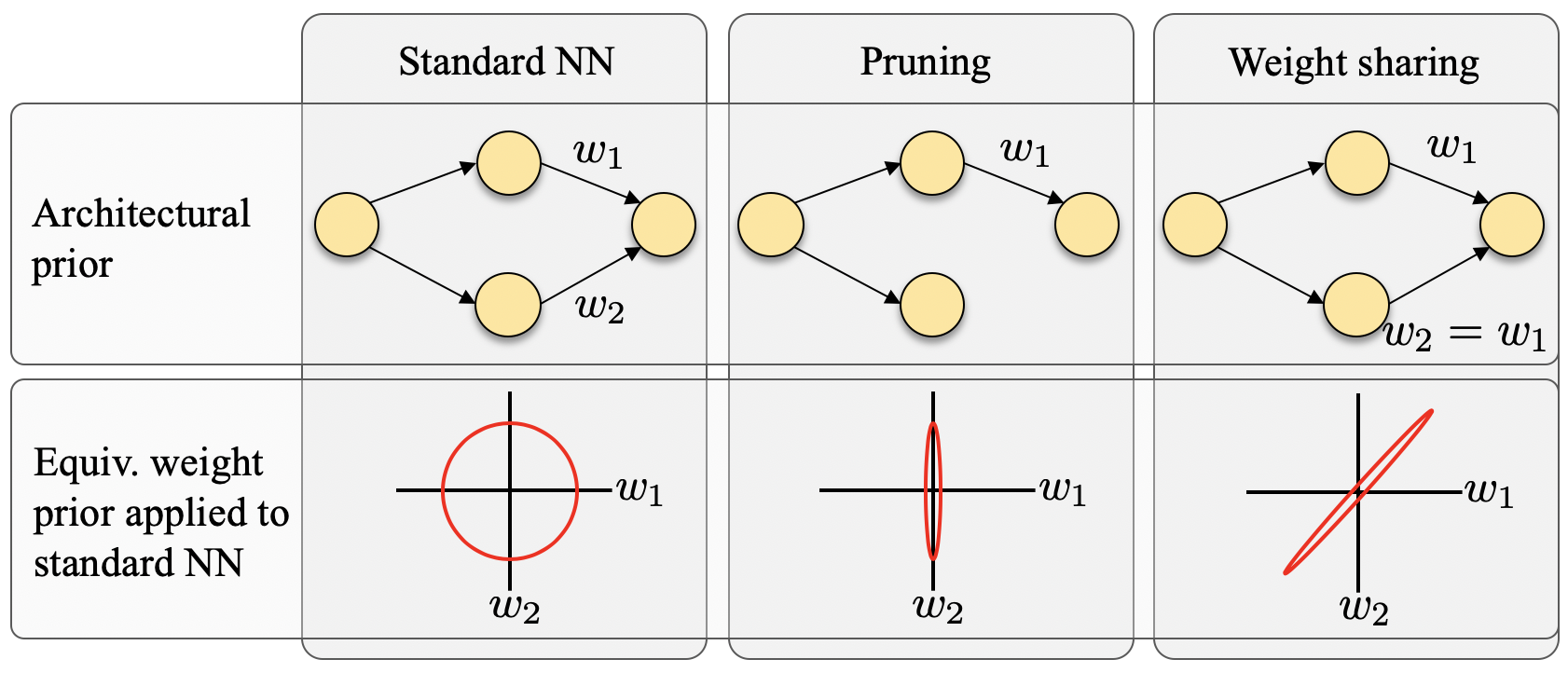}}
\caption{Weight and architectural priors achieve similar effects.}
\label{fig_weight_arch_equiv}
\end{center}
\vskip -0.2in
\end{figure}

Within an architectural prior, specific weight values of the NN determine a specific input to output mapping. A `\textbf{weight prior}' refers to a probability distribution over weights (and biases) that induces a probability distribution over input to output mappings. These mappings fall within the space allowed by the architectural prior. A weight prior can achieve similar effects to an architectural prior (see figure \ref{fig_weight_arch_equiv}), while being less rigid. 


In this paper we talk about weight priors from two perspectives: 1) In the strictly Bayesian sense, with the view to specifying a prior distribution over which Bayesian inference will performed. 2) In a looser, more general sense, for regular (non-Bayesian) NNs, where a weight prior is the initialising distribution, or is enforced through regularisation (e.g. L2 weight decay corresponds to MAP inference with i.i.d.~Gaussian priors).

A major research area in deep learning has focused on discovering useful architectural priors. This has resulted in a variety of effective layer types such as convolutions, recurrence and attention, which can be composed into models such as ResNets \citep{He2016} and transformers \citep{transformer2017}. On the other hand, weight priors are generally left vague within these: In Bayesian NNs it is common to choose independent Gaussian weight priors with variance scaled by the number of incoming connections, $n_\text{in}$ \citep{Neal1997}, so for parameter $i$ in layer $l$, $\theta_i \sim \mathcal{N}(0, \alpha/n_{\text{in},l})$, for some $\alpha$. Popular initialising schemes operate similarly, drawing weights in each layer from $\mathcal{N}(0, 2/n_\text{in})$ (He \citeyear{He2016}) or $\mathcal{N}(0, 2/(n_\text{in} + n_\text{out}))$ (Glorot \citeyear{Glorot2010}). We will refer to these as `i.i.d.~priors', since weights are i.i.d.~within layers.

\begin{figure*}[t]
\vskip 0.05in
\begin{center}

\hspace{0.6in}
\includegraphics[width=0.42\columnwidth]{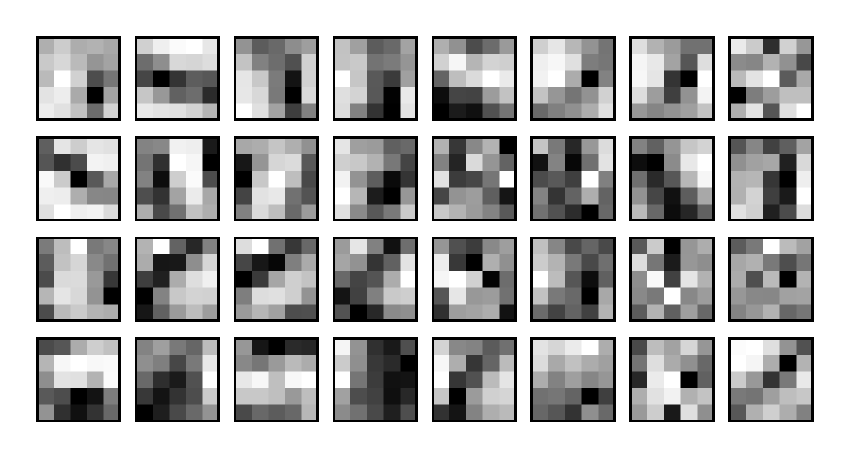}
\put(-132,25){\small LeNet5}
\put(-132,15){\small MNIST}
\put(-132,5){\small $5\times5$}
\put(-76,54){\small A) Learnt Filters}
\includegraphics[width=0.42\columnwidth]{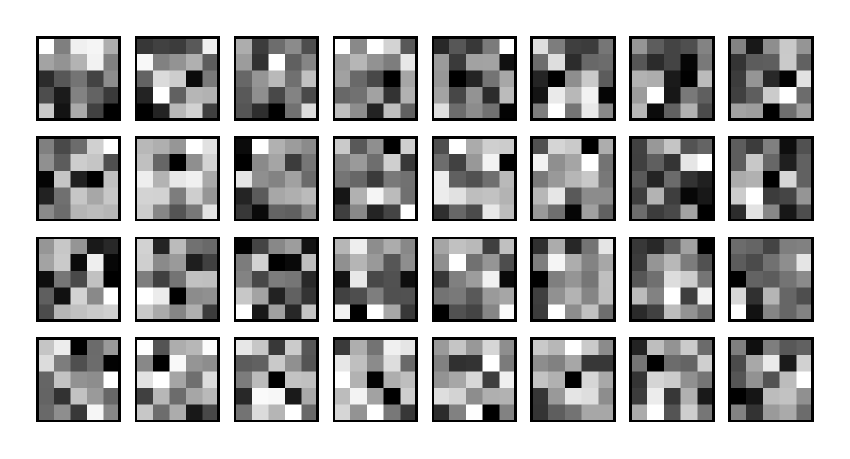}
\put(-86,54){\small B) i.i.d~Weight Prior}
\includegraphics[width=0.42\columnwidth]{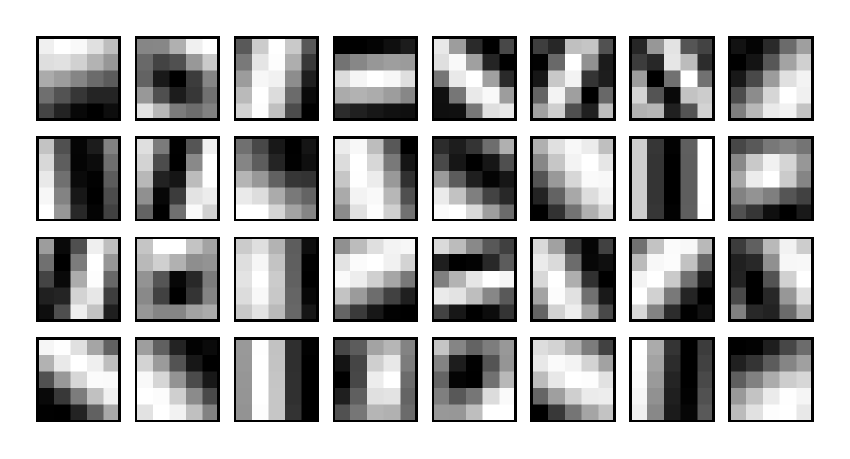}
\put(-70,64){\small Structured Prob. Gabor Weight Prior}
\put(-71,54){\small C) $\sigma_g=0.0$}
\includegraphics[width=0.42\columnwidth]{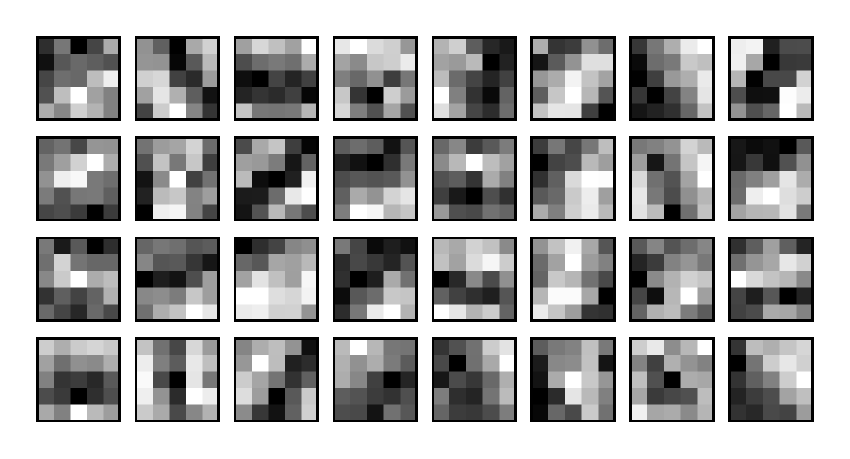}
\put(-71,54){\small D) $\sigma_g=0.3$}

\hspace{0.6in}
\includegraphics[width=0.415\columnwidth]{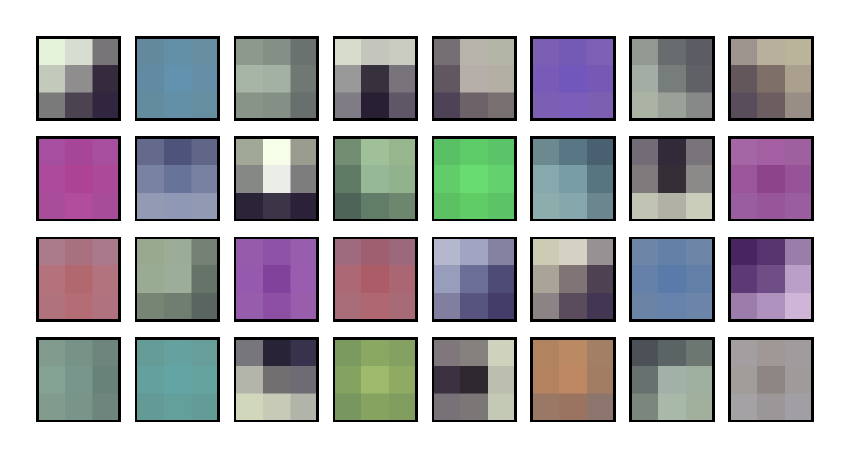}
\put(-132,25){\small VGG16}
\put(-132,15){\small ImageNet}
\put(-132,5){\small $3\times3$}
\includegraphics[width=0.415\columnwidth]{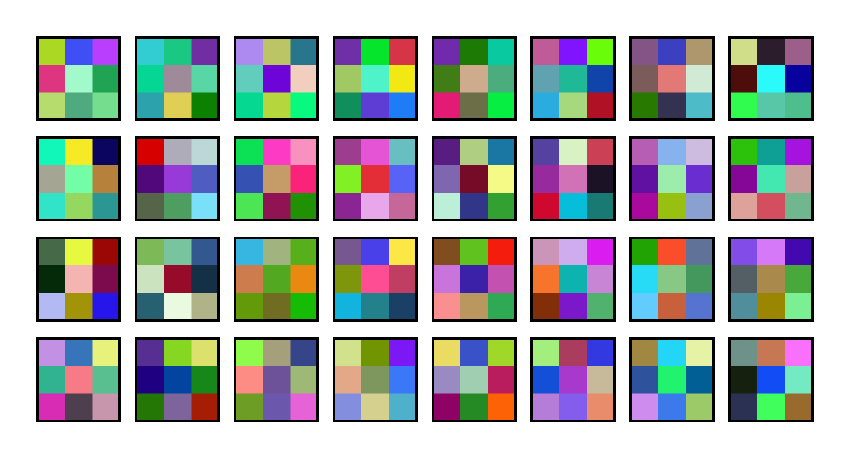}
\includegraphics[width=0.415\columnwidth]{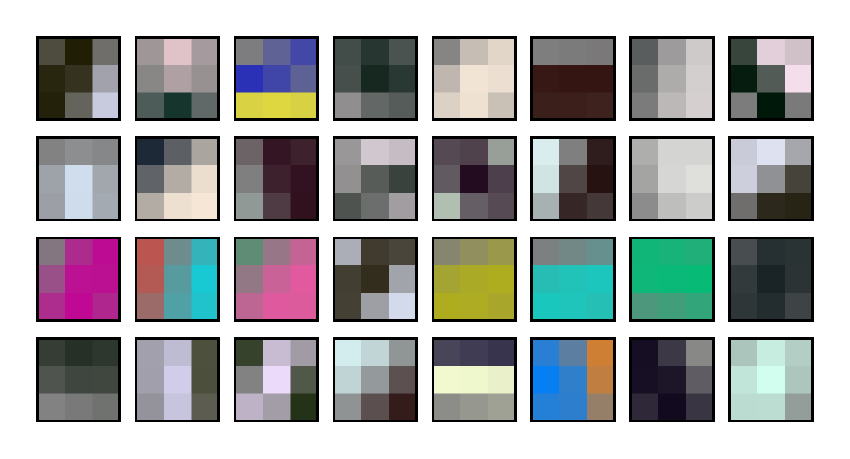}
\includegraphics[width=0.415\columnwidth]{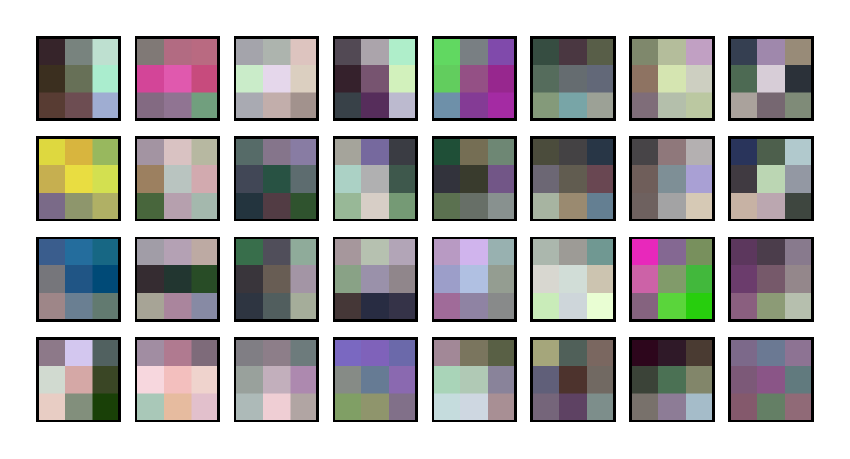}

\hspace{0.6in}
\includegraphics[width=0.415\columnwidth]{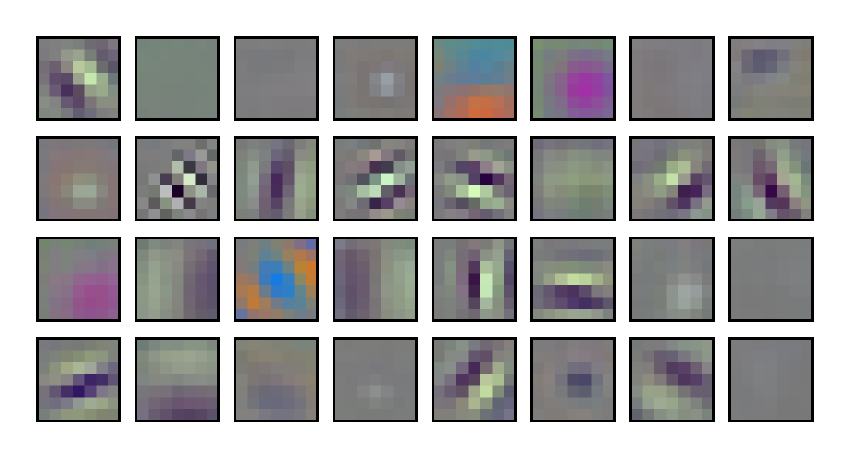}
\put(-132,25){\small ResNet50}
\put(-132,15){\small ImageNet}
\put(-132,5){\small $7\times7$}
\includegraphics[width=0.415\columnwidth]{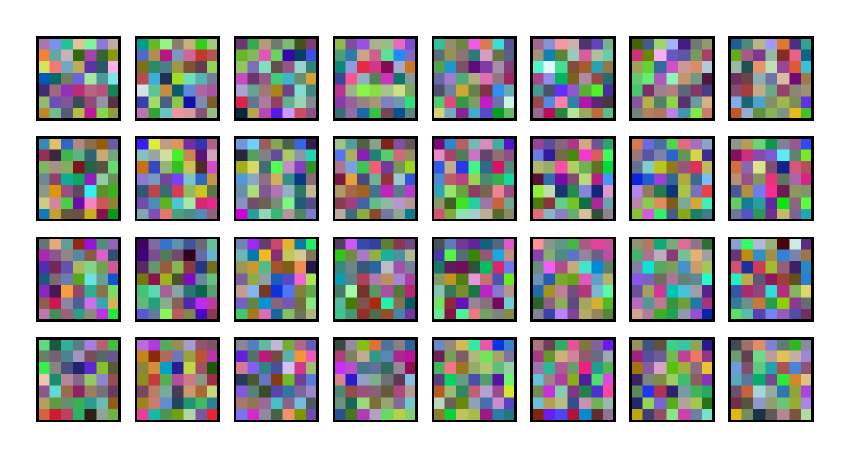}
\includegraphics[width=0.415\columnwidth]{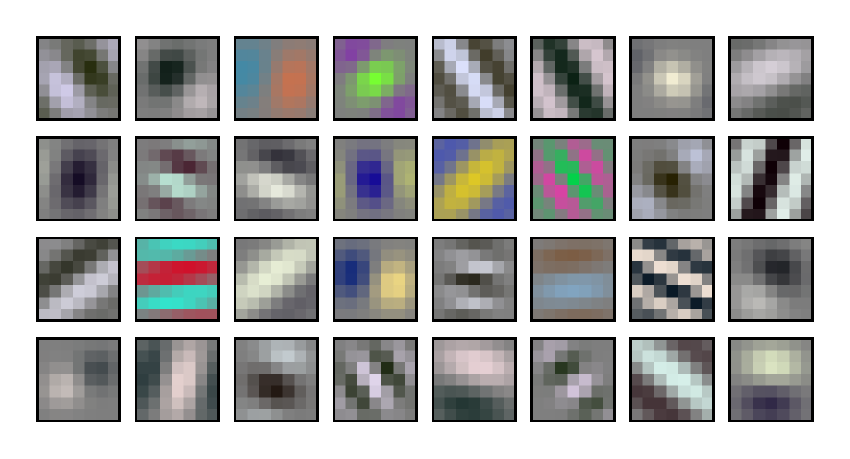}
\includegraphics[width=0.415\columnwidth]{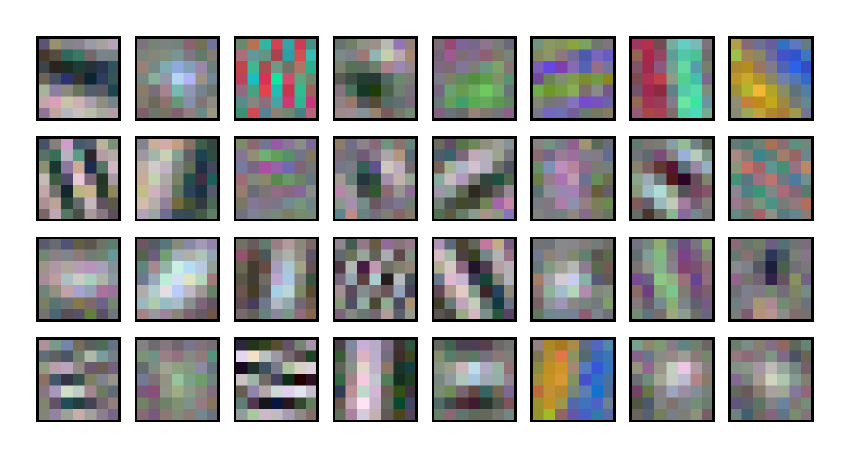}

\vskip -0.1in
\caption{A) Learnt first-layer convolutional filters for various tasks, filter sizes, and NN architectures. B) The i.i.d.~weight priors these filters were initialised from. C) \& D) Draws from the structured weight prior proposed in this paper, noiseless and noisy versions.}
\label{fig_filter_egs}
\end{center}
\vskip -0.2in
\end{figure*}

These i.i.d.~priors are largely used out of convenience - the relationship between weight-space and function-space is complex, which makes specifying a more meaningful weight prior challenging. This paper explores ways this might be done. It proposes a prior for first-layer convolutional filters based on Gabor filters (section \ref{sec_first_layer_conv}), and a method for specifying final-layer weight priors (section \ref{sec_final_layer_ws}). 

Empirical tests assessing prior quality suggest this Gabor prior is favourable compared to i.i.d.~priors (section \ref{sec_exp}). Furthermore, directly using these structured weight priors as initialising distributions for non-Bayesian NNs can deliver small but consistent performance gains in training speed and final accuracy. We believe these results are of high interest to the ongoing debate on the appropriateness of Bayesian NN weight priors, discussed in section \ref{sec_discuss}.

\section{First-Layer Convolutional Filters}
\label{sec_first_layer_conv}

When trained on vision tasks, first-layer filters of a convolutional NN learn structured weight patterns allowing the detection of low level properties such as edges and colours \citep{alexnet2012}. Examples of learnt filters from LeNet \citep{LeCun1989}, VGG \citep{Simonyan2014} and ResNet \citep{He2016}, on various datasets, are visualised in figure \ref{fig_filter_egs} A).

The figure shows a stark difference between the structures learnt and the i.i.d.~weight priors they were initialised from. Whilst the weight prior is vague enough that its support contains the learnt filters (Gaussian white noise supports any image), we now design a \emph{structured} weight prior capturing properties of the learnt filters directly. We leverage the observation that the learnt filters of convolutional NNs mimic the early processing stages in animal visual systems (V1 region), which are well approximated by Gabor filters \citep{Mardelja1980}.

\subsection{Structured Probabilistic Gabor Weight Priors}

This section shows how Gabor filters can be specified as an implicit probability distribution for the first layer filters of a convolutional NN. We model each filter to be independent of the others, although dependence could easily be introduced. Consider the real part of the Gabor function,
\begin{equation}
\label{eq_gabor}
g(f_x,f_y) = \exp{ \left( -\frac{x_\theta^2  + \gamma y_\theta^2  }{2 \sigma^2} \right)}  \cos\left(\frac{2\pi x_\theta}{ \lambda} + \psi\right)
\end{equation}
where $f_x,f_y \in \{1,2...f_w\}$ specify coordinates within the filter, $f_w$ is the filter width, and,
\begin{align}
x_\theta &= f_x \cos(\theta_g) + f_y \sin(\theta_g) \\
y_\theta &= -f_x  \sin(\theta_g) + f_y \cos(\theta_g).
\end{align}
It is straightforward to convert this to an implicit distribution by placing a probability distribution over the parameters controlling the Gabor function. Following some tuning, this work used the uniform distributions, $
\theta_g \sim \mathcal{U}(0,\pi), 
\sigma \sim \mathcal{U}(2,10), 
\lambda \sim \mathcal{U}(1,f_w),
\psi \sim \mathcal{U}(-\pi,\pi), 
\gamma \sim \mathcal{U}(0,1.5)$.

The function $g(f_x,f_y)$ applies to single channels, and can be used directly with grayscale inputs. Applying them to RGB inputs is more challenging - Palm et al. (\citeyear{Palm2000}) summarise several approaches for this. We propose something similar to what they describe as `unichrome features'. Our modified version is given in algorithm \ref{alg_rgb_gabor}, explicitly sampling black \& white filters with a certain probability.

The approach described above samples filters that \emph{strictly} follow a Gabor function. 
A simple way to relax this constraint is to add independent $\mathcal{N}(0, \sigma_g^2)$ noise to $g(f_x, f_y)$.
%
Section \ref{sec_exp}'s experiments are run using $\sigma^2_g=0$ unless otherwise stated. 
Due to the implicit nature of this distribution, its use may be challenging when performing posterior inference.
In this paper, we consider primarily its quality as a Bayesian prior/initialising distribution, and leave questions of inference to future work. It is possible that inference could be performed by fitting the Gabor parameters and $\sigma^2_g$ using maximum likelihood.
Figure \ref{fig_filter_egs} C) \& D) show examples of prior draws from grayscale and colour versions of this Gabor prior, for varying filter sizes and $\sigma^2_g$. Note their similarity with the learnt filters.


\section{Final-Layer Weights}
\label{sec_final_layer_ws}

We next propose a method for specifying a structured weight prior for the weights in the final, fully-connected layer of a classifier NN.
Let there be $n$ total parameters in the NN, forming a flattened vector $\theta := (\theta_{1},\theta_{2}, \hdots, \theta_{n})$. If the last $m$ of these are final-layer weights, we define $\theta_\text{final} := (\theta_{n-m + 1}, \hdots, \theta_{n} )$, and, $\theta_{\lnot \text{final}} := (\theta_{1}, \hdots, \theta_{n-m} )$.
Originally we trialled schemes adding structure within $P(\theta_{\text{final}})$, but independently of earlier initialisations in the NN, $P(\theta)=P(\theta_{\lnot \text{final}}) P(\theta_{\text{final}})$. This approach had little success. Our experience suggests that a more promising approach is to condition weight priors later in the NN, on values drawn earlier,
$P(\theta)=P(\theta_{\lnot \text{final}}) P(\theta_{ \text{final}}|\theta_{\lnot \text{final}})$. 
Intuitively, this is because it is necessary to know what a \emph{specific} hidden feature represents before assigning a prior to its outgoing weights.

\begin{table}[t]
\caption{Mean entropy of prior predictive distribution for CNNs on training set. Standard errors are of order $0.02$.}
\label{tab_entropies}
\vskip -0.2in
\begin{center}
\resizebox{1.\columnwidth}{!}{
\begin{small}
\begin{tabular}{l | cc | cc | cc }
\toprule
\multicolumn{1}{c}{}  &\multicolumn{2}{c}{-- 1 layer --} & \multicolumn{2}{c}{-- 2 layer --} & \multicolumn{2}{c}{-- 3 layer --}\\
& \scriptsize{CNN iid} & \scriptsize{CNN Gabor} 
& \scriptsize{CNN iid} & \scriptsize{CNN Gabor} 
& \scriptsize{CNN iid} & \scriptsize{CNN Gabor} \\
\midrule
MNIST  & 1.24 & 1.40 & 1.12 & 1.29 & 1.01 & 1.14 \\
Fashion MNIST  & 1.04 & 1.34 & 0.94 & 1.19 & 0.87 & 1.08 \\
CIFAR 10  & 0.75 & 1.05 & 0.73 & 0.93 & 0.69 & 0.89 \\
\bottomrule
\end{tabular}
\end{small}
}
\end{center}
\vskip -0.1in
\end{table}

\begin{figure}[t]
\vskip 0.05in
\begin{center}
\small{5$\times$ i.i.d.~priors}

\includegraphics[width=0.19\columnwidth]{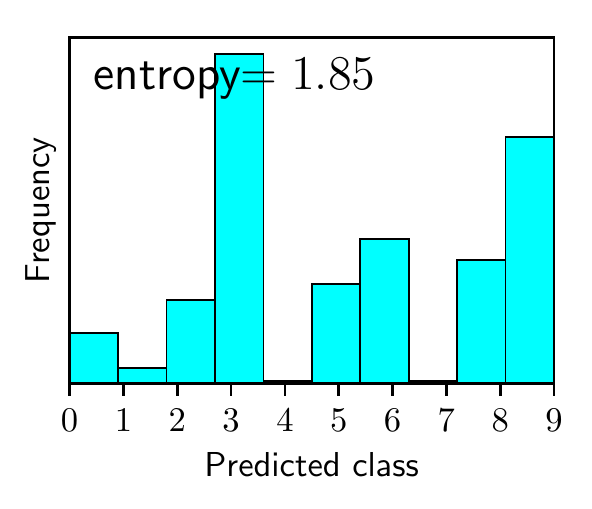}
\includegraphics[width=0.19\columnwidth]{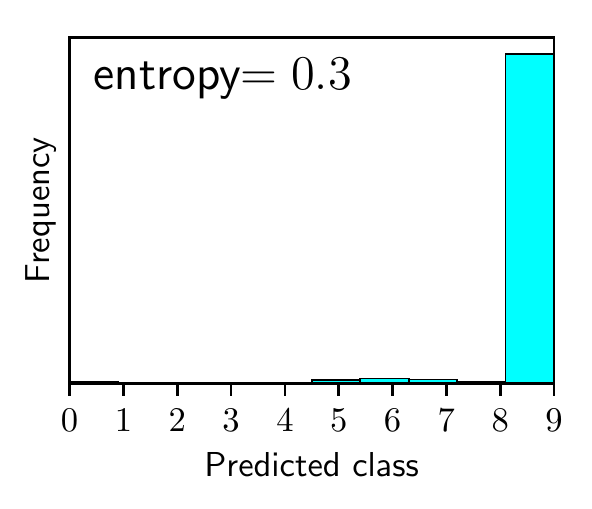}
\includegraphics[width=0.19\columnwidth]{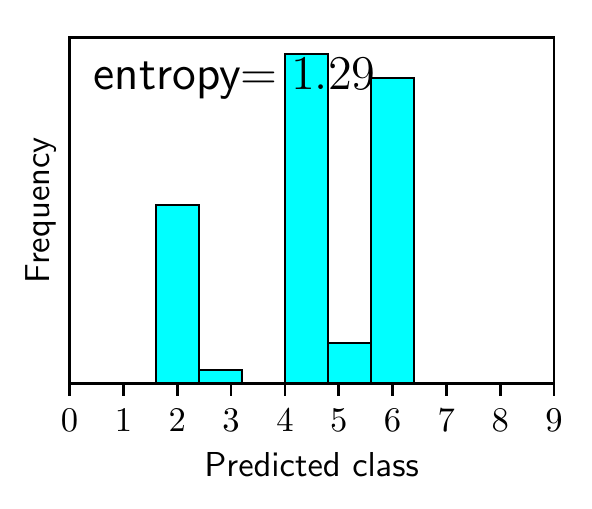}
\includegraphics[width=0.19\columnwidth]{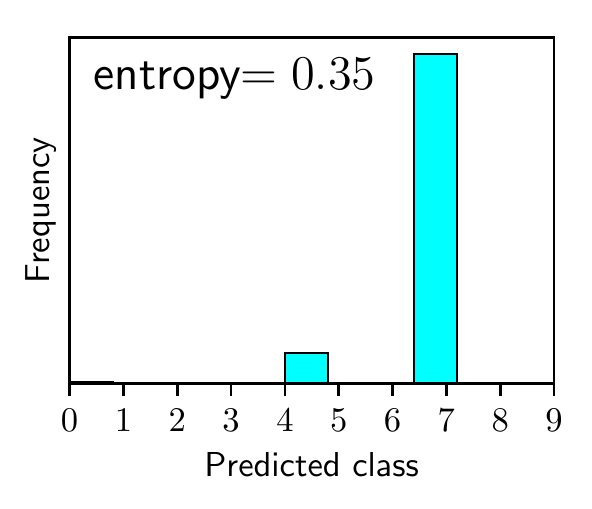}
\includegraphics[width=0.19\columnwidth]{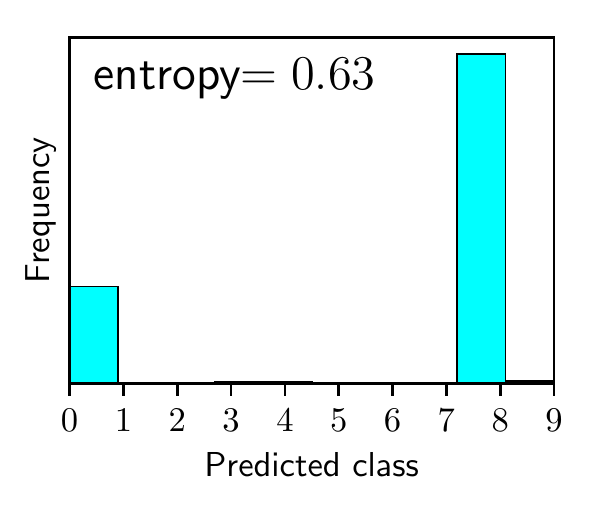}

\small{5$\times$ structured Gabor priors}

\includegraphics[width=0.19\columnwidth]{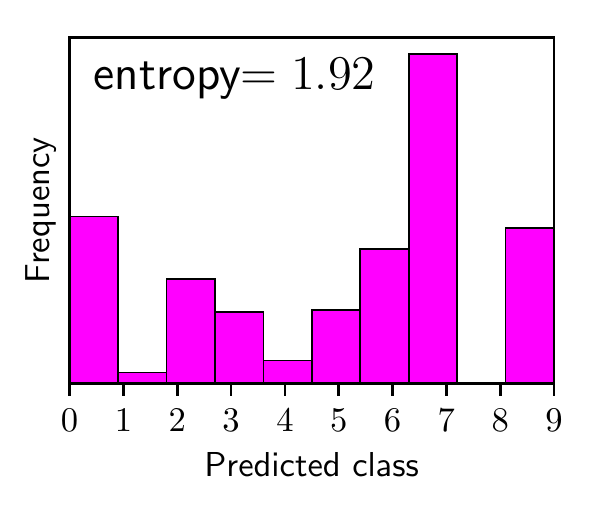}
\includegraphics[width=0.19\columnwidth]{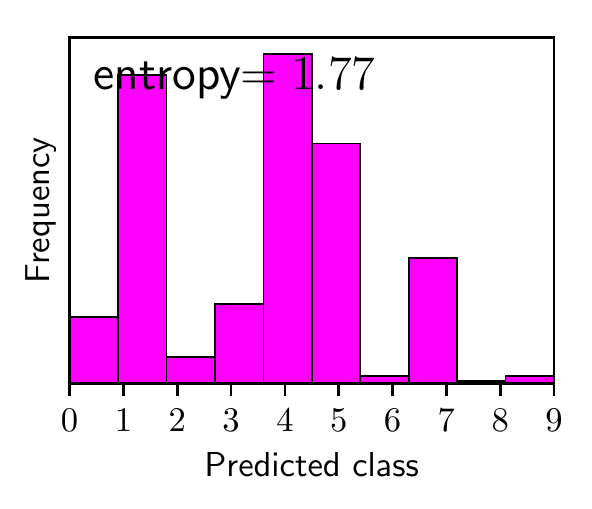}
\includegraphics[width=0.19\columnwidth]{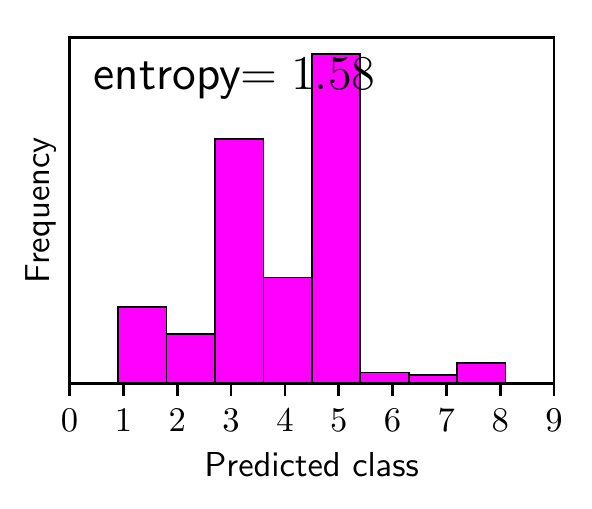}
\includegraphics[width=0.19\columnwidth]{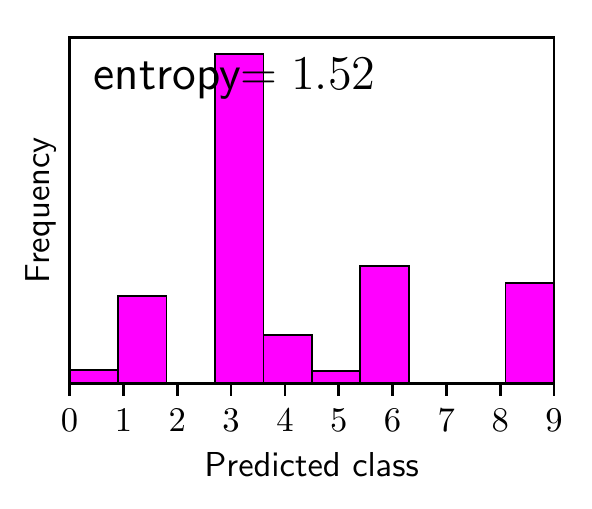}
\includegraphics[width=0.19\columnwidth]{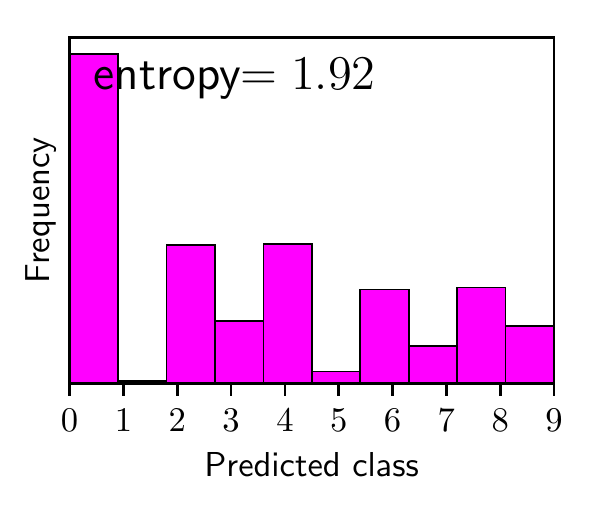}

\vskip -0.1in
\caption{Example histograms of predictions made by CNN prior samples on the MNIST training set, and entropy of the distribution.}
\label{fig_entropy_egs}
\end{center}
\vskip -0.2in
\end{figure}

\subsection{Feature Specific Priors}
\label{sec_final_feats}

Specifying conditional priors is challenging, and this method uses an empirical, heuristic mechanism, described in algorithm \ref{alg_feats}. 
The key idea is to sample a small number (e.g.~10 or 20) inputs from each class, and record how `activated' each final-layer hidden node is by each class. These activation statistics can be used to sensibly structure the final layer weight prior.

For example, consider a NN with two outputs predicting two classes (A and B). If the $i^\text{th}$ final-layer hidden node is strongly activated by examples of class A, but not class B, the weight connecting hidden node $i$ to output A should be large and positive, whilst the connection to class B should be smaller, or even negative. Note that it is the \textit{relative} difference between activations that is important.
This prior performs a mild form of empirical Bayes, as it is specified using (a small) amount of the training data.

\section{Experiments}
\label{sec_exp}

The main objective of this section is to empirically assess whether the structured weight priors proposed in this paper can provide advantages over i.i.d.~weight priors for convolutional NNs (CNNs). We further include some comparisons with fully-connected (fc) NNs, also with i.i.d.~weight priors, to allow observations of the effect of architectural vs weight priors.
Testing might ideally be done through full computation of the Bayesian posterior.
However, this is challenging for our Gabor prior (since it is an implicit distribution), and risks conflating characteristics of the prior with artefacts of the inference process. Instead, we resort to more direct ways of assessing prior quality. Four tests are used.

1) \textbf{Prior predictive diversity} - tests general sensitivity to different parts of the training distribution. Adapted from \citep{Wenzel2020}.

2) \textbf{Prior activation correlations} - tests whether more similar inputs give rise to more similar responses. Adapted from \citep{Wilson2020}. 

3) \textbf{Class-Agnostic Prior Predictive Accuracy (CAPPA)} - tests how useful the class decision boundaries induced by the prior are for a classification task.

4) \textbf{Training curves} - allows some (approximate) measure of the `distance' between the prior and posterior (convergence rate), and whether the prior helps find a better final solution (final test accuracy).


Tests 1) 2) \& 3) are used with only the structured Gabor prior, whilst 4) is used to test both the Gabor prior and feature specific prior of section \ref{sec_final_feats}.
We perform these experiments on NN architectures and datasets that are small by modern standards (one to three layers). This is because we only structure priors in the first and final layers, and adding layers in between decreases the observable effect of our method.

The i.i.d.~weight priors compared against (both fc NN and CNN) draw weights for each layer from $\mathcal{N}(0,2/n_\text{in})$ \citep{Neal1997,He2016}, where $n_{\mathrm{in}}$ is the number of inputs to each node. Biases were initialised to zero. CNNs using the structured weight prior also use this i.i.d.~prior for weights in layers that are not structured. For comparability, after sampling from the Gabor prior, we apply a postprocessing step so that the mean and variance of the whole layer matches that of the i.i.d.~prior. The CNNs used have $16$ first layer filters of size $5 \times 5$, followed by a maxpooling layer. Subsequent layers in the CNN have $16^l$ filters of size $3 \times 3$ for layer $l$. The fc NNs have $1024$ hidden units per layer.

\subsection{Prior Predictive Diversity}
\label{sec_exp_predictive_diversity}

Predictions for the training set are made using an untrained NN, initialised from the prior distribution. For each data point, the class with highest predictive probability is recorded, and a histogram of these are compiled - see figure \ref{fig_entropy_egs} for five examples. The entropy of this histogram measures how spread out the prior predictions are between classes. A good prior should have high entropy.

Table \ref{tab_entropies} records the mean entropy of 500 histograms for CNNs of varying depth, using i.i.d.~and structured Gabor priors from section \ref{sec_first_layer_conv}. In general, the Gabor prior produces distributions with higher entropies, suggesting it is more sensitive to the input distribution, and potentially a more useful prior.


\begin{table}[t]
\caption{Activation correlations for pairs of inputs, when inputs are drawn from the same class vs.~different classes.}
\label{tab_activ_corr}
\vskip 0.05in
\begin{center}
\resizebox{1.\columnwidth}{!}{
\begin{small}
\begin{tabular}{l | ccc | ccc | ccc }
\toprule
\multicolumn{1}{c}{}  &\multicolumn{3}{c}{-- 1 layer --} & \multicolumn{3}{c}{-- 2 layer --} & \multicolumn{3}{c}{-- 3 layer --}\\
& \tiny{fcNN iid} & \tiny{CNN iid} & \tiny{CNN Gabor} 
& \tiny{fcNN iid} & \tiny{CNN iid} & \tiny{CNN Gabor} 
& \tiny{fcNN iid} & \tiny{CNN iid} & \tiny{CNN Gabor}  \\
\midrule
\multicolumn{1}{c}{}  &\multicolumn{9}{c}{MNIST}\\
Corr. same class 
& 0.50 & 0.76 & 0.70 
& 0.72 & 0.83 & 0.74
& 0.76 & 0.82 & 0.79   \\
Corr. diff. class 
& 0.40 & 0.66 & 0.58 
& 0.66 & 0.76 & 0.65 
& 0.71 & 0.74 & 0.72   \\
\multicolumn{1}{c}{}  &\multicolumn{9}{c}{Fashion MNIST}\\
Corr. same class  
& 0.84 & 0.86 & 0.81 
& 0.84 & 0.90 & 0.79 
& 0.84 & 0.92 & 0.82   \\
Corr. diff. class 
& 0.72 & 0.69 & 0.65 
& 0.74 & 0.81 & 0.61 
& 0.76 & 0.84 & 0.64   \\
\multicolumn{1}{c}{}  &\multicolumn{9}{c}{CIFAR 10}\\
Corr. same class  
& 0.86 & 0.89 & 0.70 
& 0.87 & 0.86 & 0.84 
& 0.92 & 0.91 & 0.83 \\
Corr. diff. class 
& 0.85 & 0.88 & 0.69 
& 0.86 & 0.85 & 0.83 
& 0.91 & 0.91 & 0.82 \\
\bottomrule
\end{tabular}
\end{small}
}
\end{center}
\vskip -0.1in
\end{table}

\begin{figure}[t]
\vskip 0.0in
\begin{center}
\includegraphics[width=0.24\columnwidth]{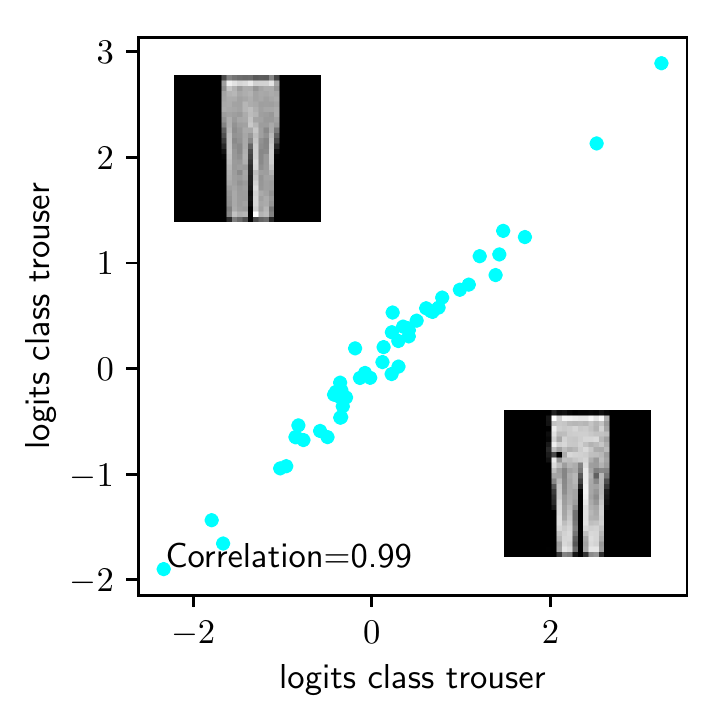}
\includegraphics[width=0.24\columnwidth]{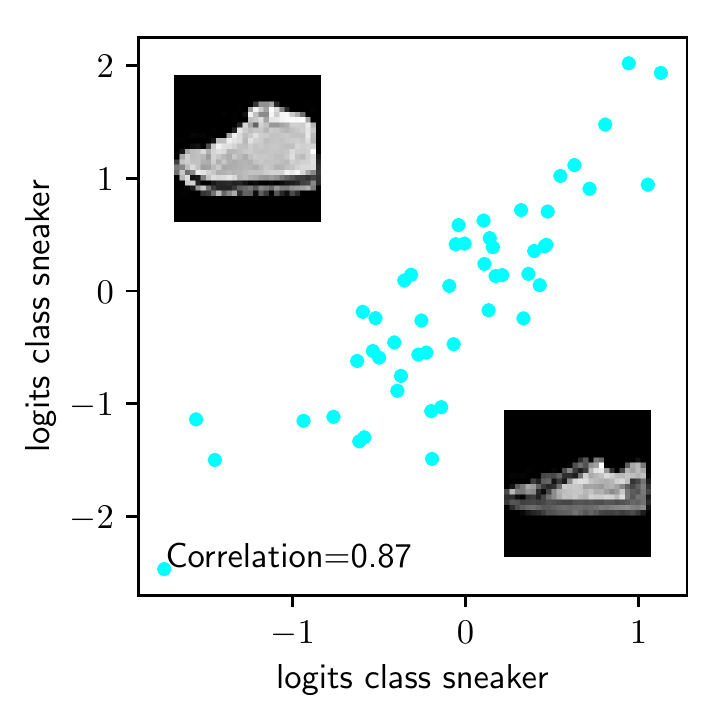}
\put(-76,55){\small -- Same class --}
\includegraphics[width=0.24\columnwidth]{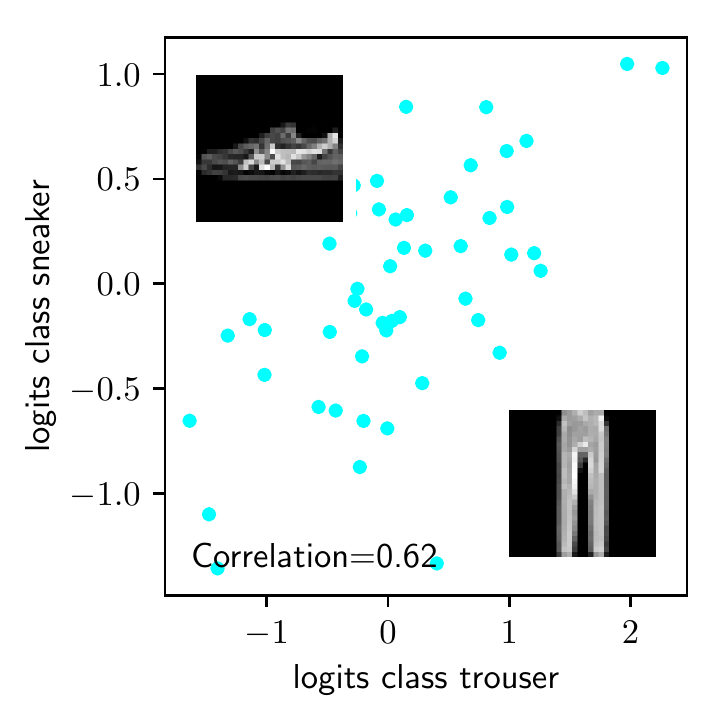}
\includegraphics[width=0.24\columnwidth]{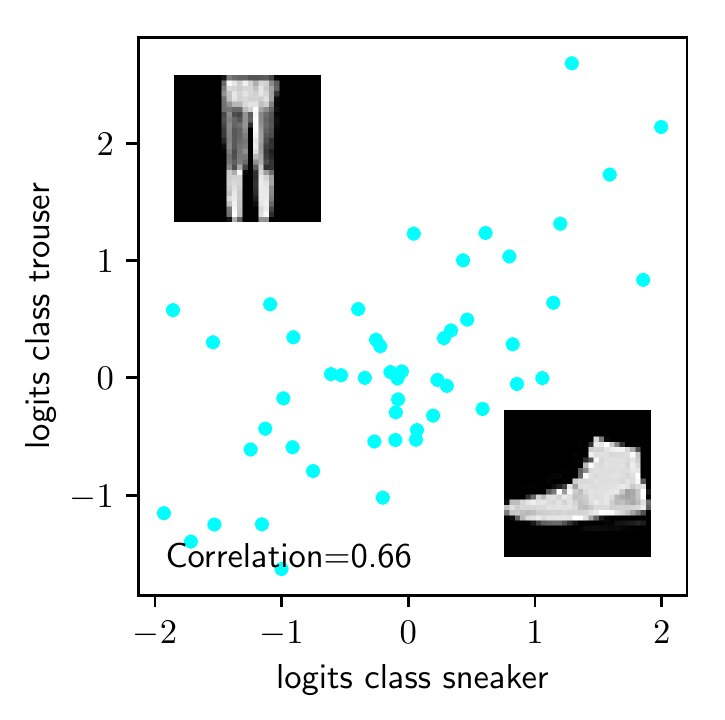}
\put(-85,55){\small -- Different classes --}
\put(-238,13){\rotatebox{90}{\small CNN i.i.d.}}

\includegraphics[width=0.24\columnwidth]{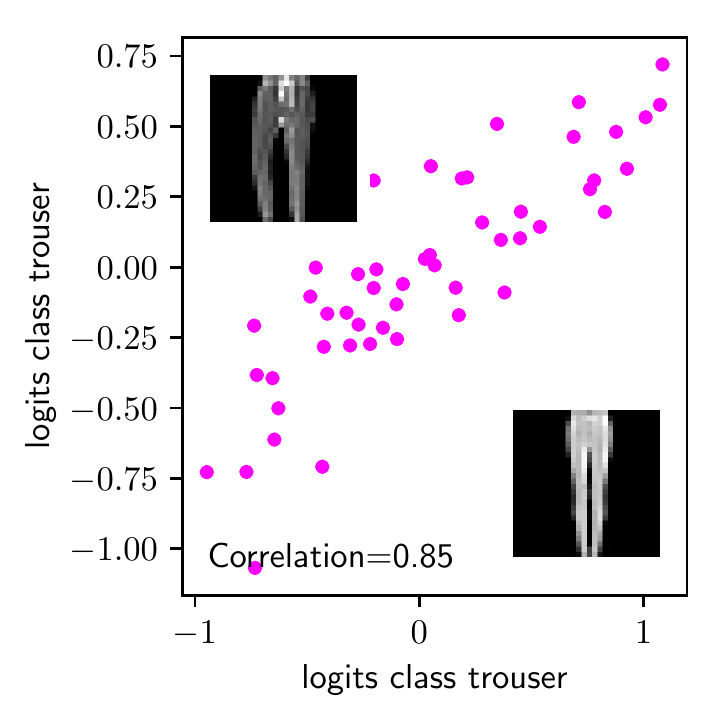}
\includegraphics[width=0.24\columnwidth]{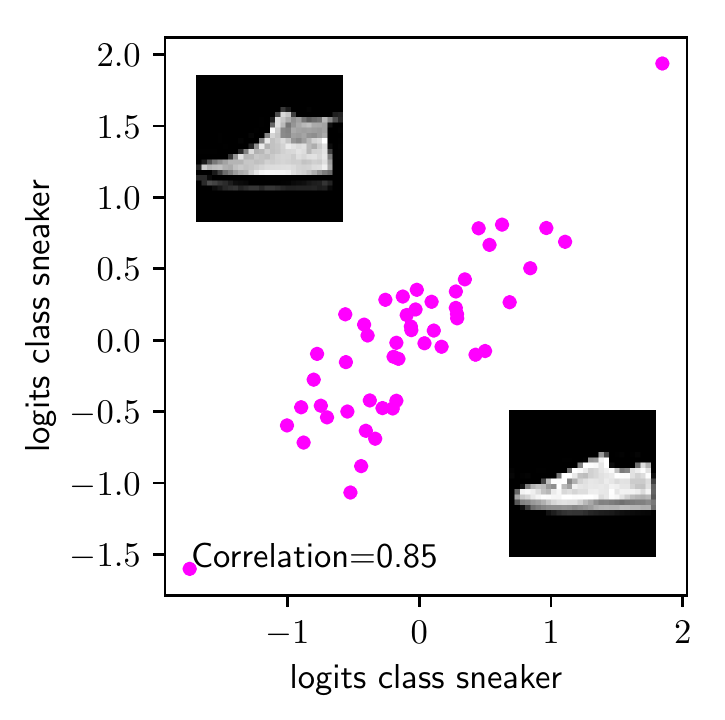}
\includegraphics[width=0.24\columnwidth]{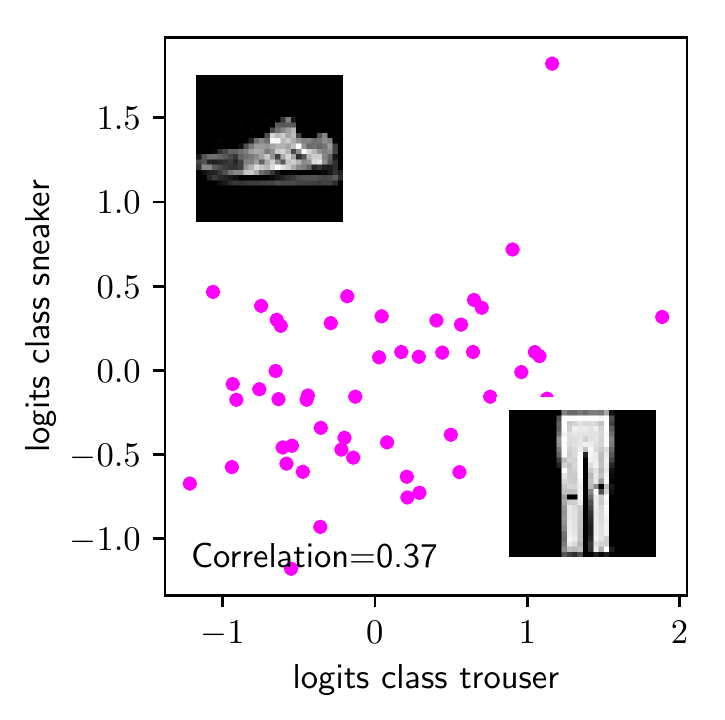}
\includegraphics[width=0.24\columnwidth]{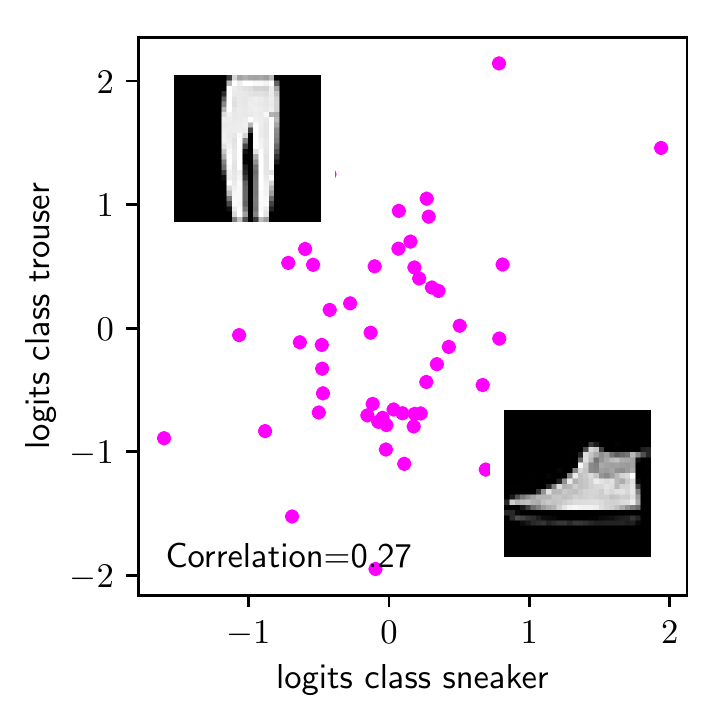}
\put(-238,10){\rotatebox{90}{\small CNN Gabor}}
\vskip -0.1in
\caption{Correlation between logits at one output for four input pairs, for 50 samples from the prior. Single-layer convolutional NN, Fashion MNIST.}
\label{fig_logits_corr_egs_small}
\end{center}
\vskip -0.2in
\end{figure}

\subsection{Prior Activation Correlations}
\label{sec_exp_act_corr}

This test records correlations at a single output of a NN for input pairs when 1) input pairs are from the same class, 2) input pairs are from different classes. A good prior should have higher correlations for inputs of the same class, and lower correlations for different classes.
Correlations are computed over multiple prior draws, which are averaged over multiple input pairs - figure \ref{fig_logits_corr_egs_small} illustrates how these correlations are calculated for several input pairs. 

Table \ref{tab_activ_corr} shows results. Correlations were computed over 50 prior draws, and averaged over 10,000 examples of each class type. In general the correlations for both same-class and different-class input pairs are lower for CNN Gabor than for CNN i.i.d.. It's often the case that the CNN Gabor prior creates the largest difference between same-class and different-class correlations, signaling that it may be a more appropriate prior, though this pattern does not always hold.



\begin{figure}[t!]
\vskip 0.1in
\begin{center}
\centerline{}
\includegraphics[width=0.32\columnwidth]{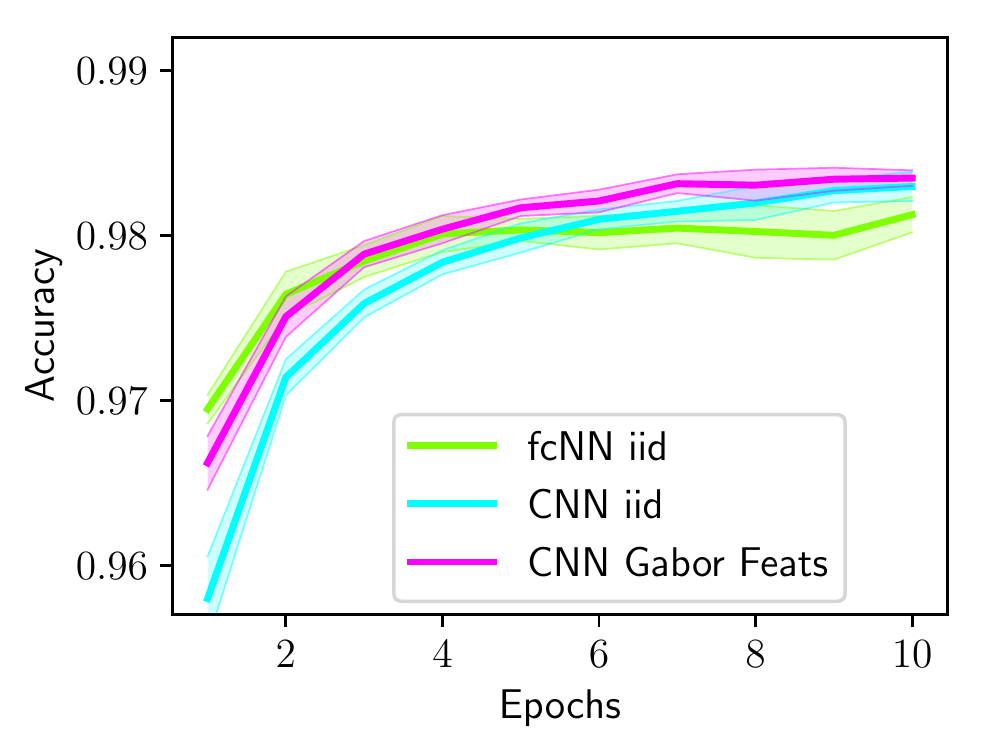}
\includegraphics[width=0.32\columnwidth]{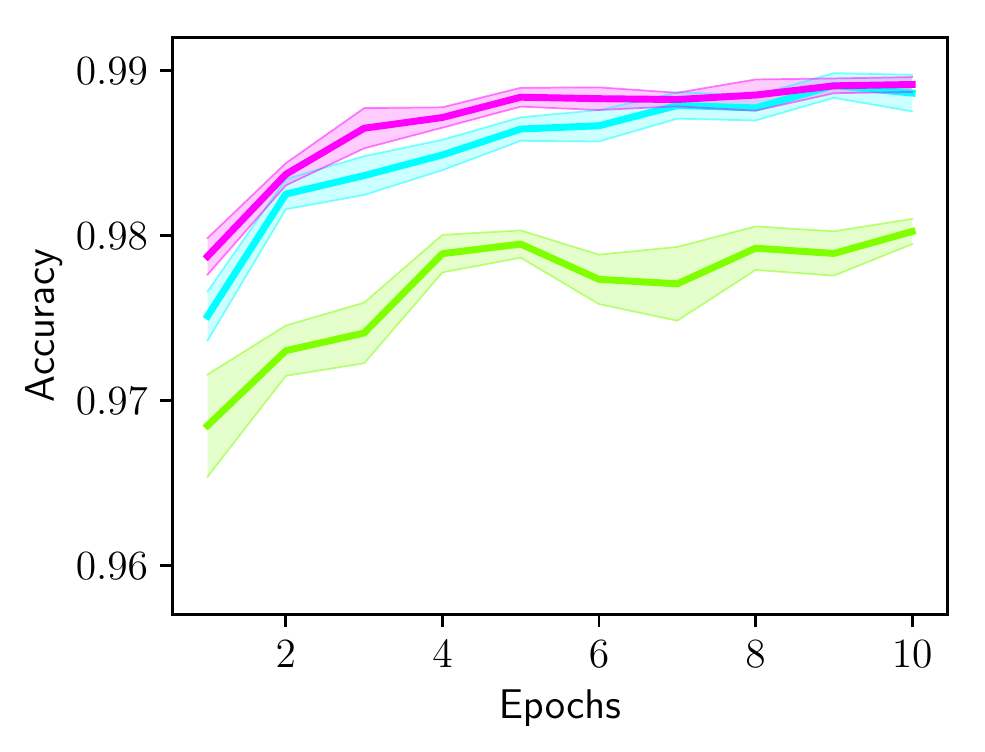}
\includegraphics[width=0.32\columnwidth]{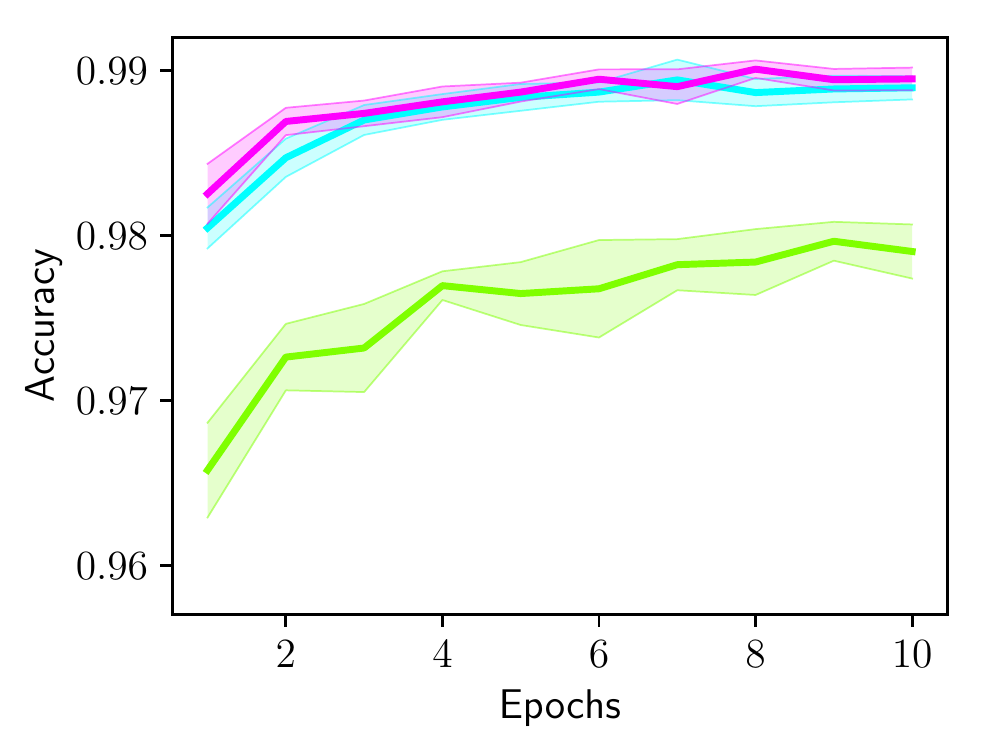}
\put(-50,62){\small 3 layer}
\put(-120,62){\small 2 layer}
\put(-200,62){\small 1 layer}
\put(-238,15){\rotatebox{90}{\small MNIST}}

\includegraphics[width=0.32\columnwidth]{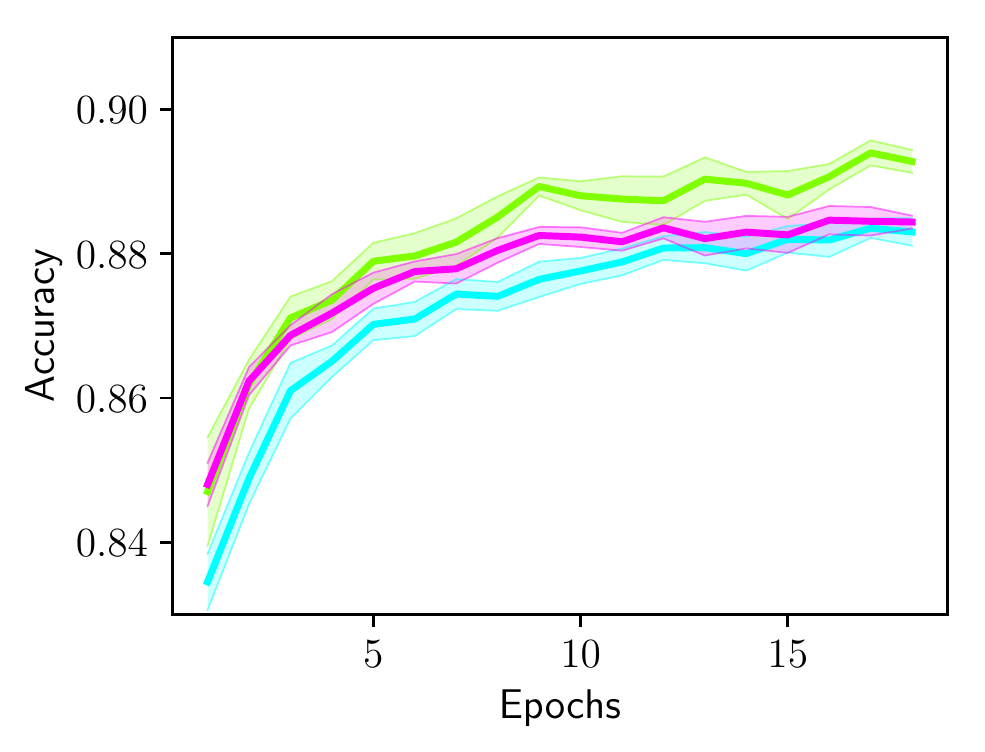}
\includegraphics[width=0.32\columnwidth]{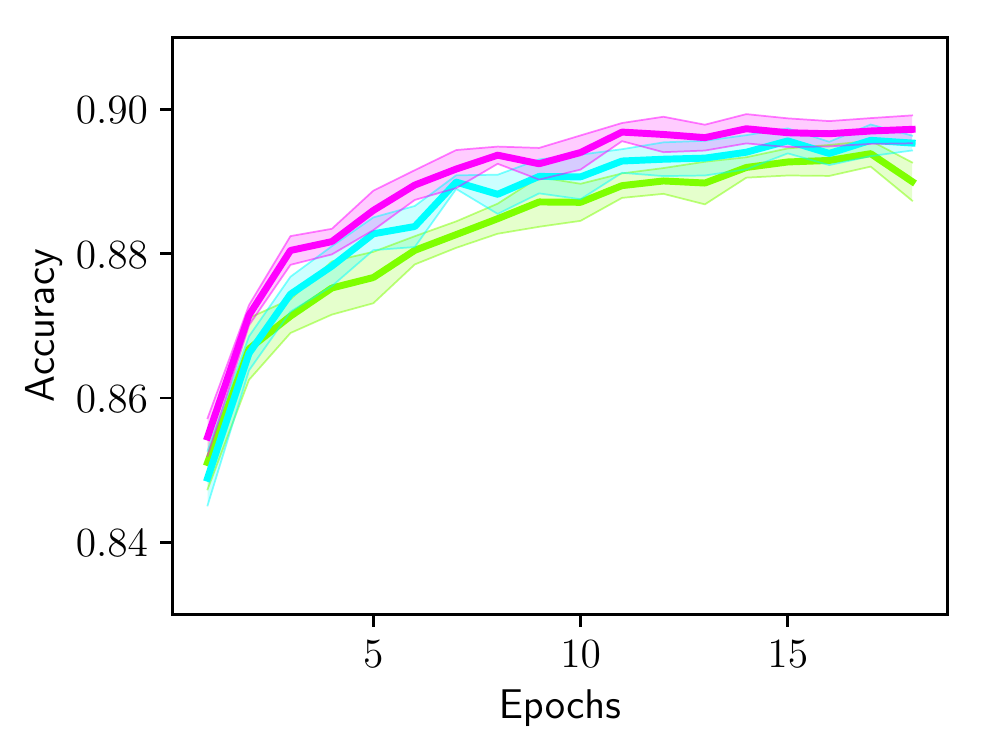}
\includegraphics[width=0.32\columnwidth]{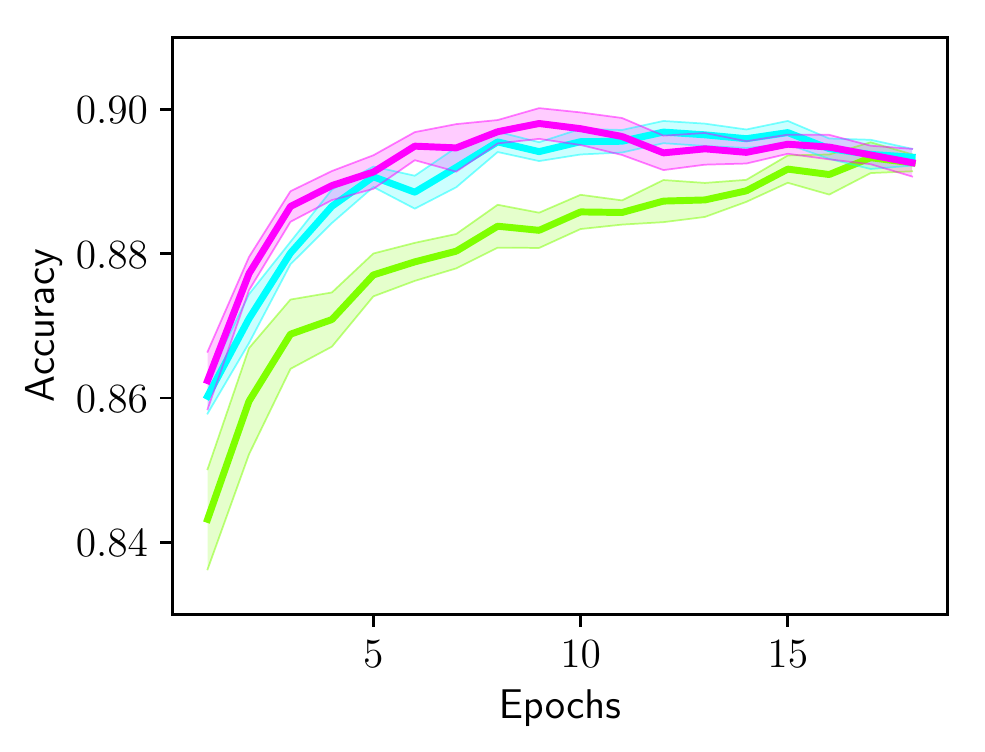}
\put(-238,15){\rotatebox{90}{\small F. MNIST}}

\includegraphics[width=0.32\columnwidth]{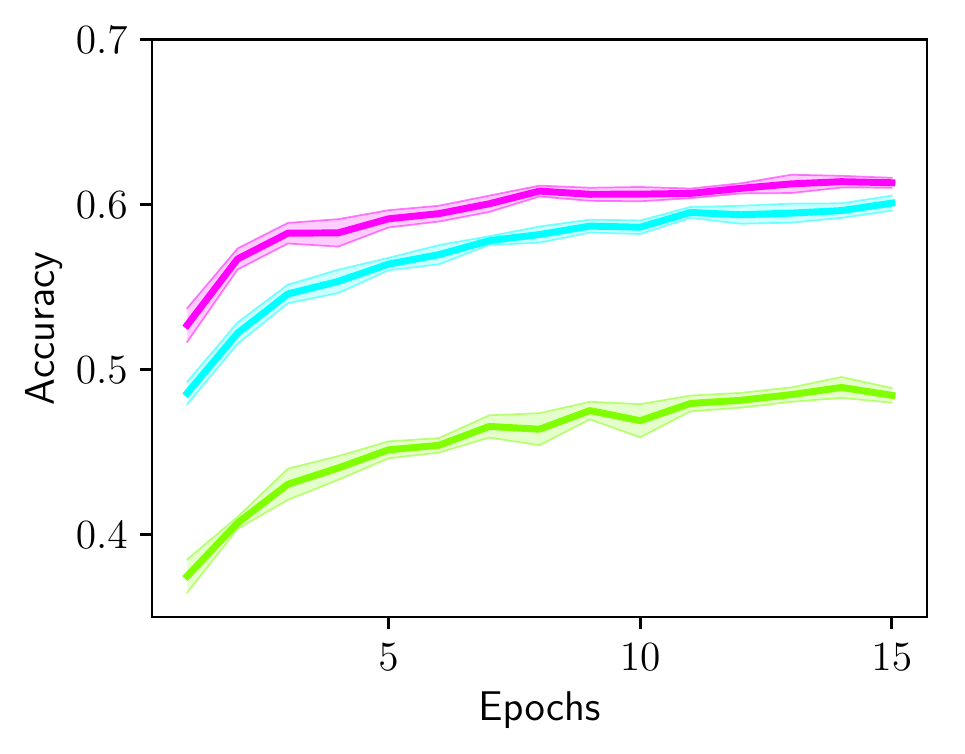}
\includegraphics[width=0.32\columnwidth]{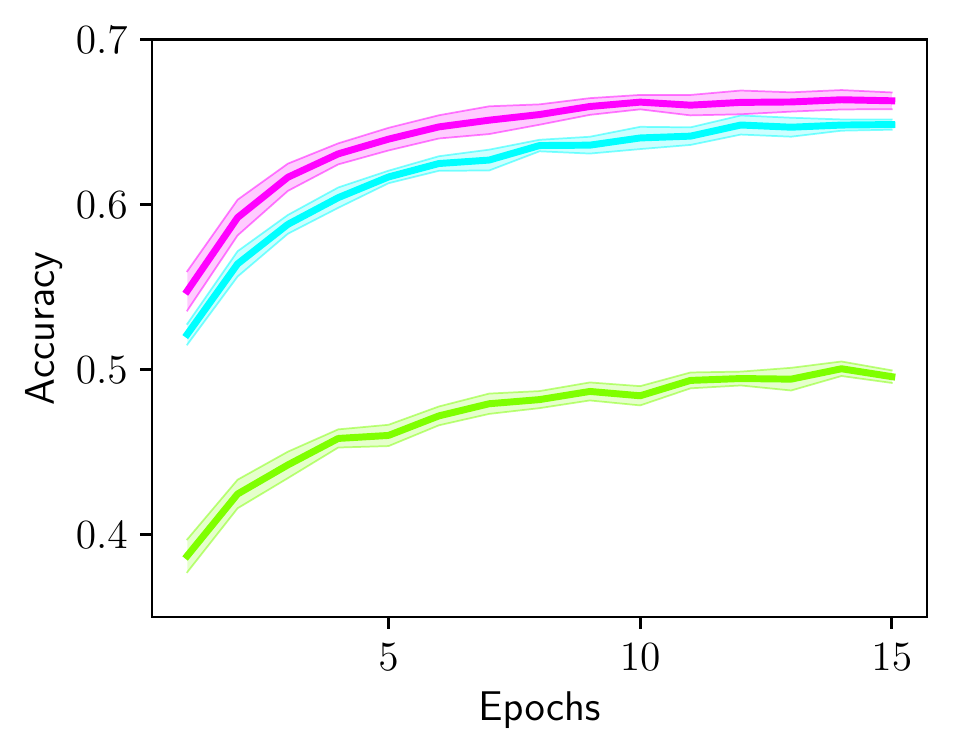}
\includegraphics[width=0.32\columnwidth]{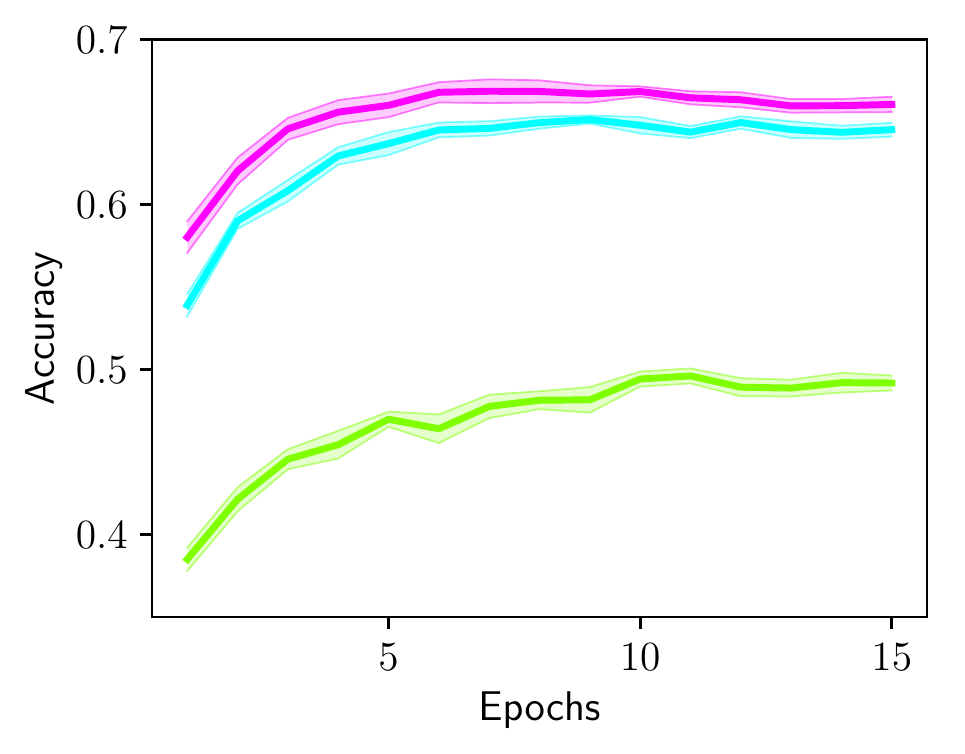}
\put(-238,15){\rotatebox{90}{\small CIFAR 10}}

\caption{Mean $\pm 2$ std. errors over ten training runs for various datasets and NN depths, with initialisations drawn from prior. For visual clarity, only test accuracy is shown.}
\label{fig_exp_train_curves}
\end{center}
\vskip -0.2in
\end{figure}

\begin{table}[b]
\vskip -0.2in
\caption{Mean CAPPA for binary classification tasks. Standard errors are of order $0.002$.} 
\label{tab_cappa}
\vskip -0.1in
\begin{center}
\resizebox{1.\columnwidth}{!}{
\begin{small}
\begin{tabular}{l | ccc | ccc | ccc }
\toprule
\multicolumn{1}{c}{}  &\multicolumn{3}{c}{-- 1 layer --} & \multicolumn{3}{c}{-- 2 layer --} & \multicolumn{3}{c}{-- 3 layer --}\\
& \scriptsize{fcNN iid} & \scriptsize{CNN iid} & \scriptsize{CNN Gabor} 
& \scriptsize{fcNN iid} & \scriptsize{CNN iid} & \scriptsize{CNN Gabor} 
& \scriptsize{fcNN iid} & \scriptsize{CNN iid} & \scriptsize{CNN Gabor} \\
\midrule
MNIST  
& 0.612 & 0.620 & 0.623 
& 0.591 & 0.608 & 0.612 
& 0.585 & 0.587 & 0.593 \\
Fashion MNIST  
& 0.574 & 0.582 & 0.593 
& 0.566 & 0.565 & 0.592 
& 0.563 & 0.562 & 0.581 \\
CIFAR 10
& 0.522 & 0.523 & 0.528 
& 0.520 & 0.521 & 0.525 
& 0.517 & 0.520 & 0.522 \\
\bottomrule
\end{tabular}
\end{small}
}
\end{center}
\vskip -0.1in
\end{table}

\subsection{Class-Agnostic Prior Predictive Accuracy}

For each dataset, we select $1000$ examples of two classes to create binary classification tasks (MNIST 0's vs 1's, Fashion MNIST trousers vs shirts, CIFAR cars vs birds). For each task, we use an untrained NN (with weights sampled from the prior) to classify the $2000$ data examples, and record the accuracy. As we are interested in how well the prior divides up the input space, rather than whether it can assign the correct classes a priori, we also invert the class predictions and recompute the accuracy. The CAPPA is defined as the maximum of these two accuracies. Hence, it can be at worst 0.50 and at best 1.00.

Results are shown in table \ref{tab_cappa} for fc NNs with i.i.d.~priors, CNNs with i.i.d.~priors, and CNNs with Gabor priors. The mean is computed over 500 prior draws. Across most datasets and NN depths the ordering is: fc NN~i.i.d. $<$ CNN~i.i.d. $<$ CNN~Gabor. 
In general CAPPA decreases with depth, suggesting that deeper NNs have worse priors - also observed in table \ref{tab_entropies}. 
%
%
This is counter-intuitive, but has been noted by previous authors \citep[sec 3.2]{Lee2018}. For NNs with i.i.d.~weight priors, increasing depth tends to produce functional priors that are increasingly uninteresting. 
This highlights i) the importance of choosing weight priors carefully with deep NNs and ii) that standard practice for SGD-trained deep NNs may not translate into good Bayesian functional priors.





\subsection{Training Curves}
\label{sec_exp_traincurve}

Following a parameter initialisation sampled from the prior, we train NNs in the usual way using the Adam optimiser. Broadly speaking, a better prior may be expected to converge faster, and have higher final test accuracy. 
%
%
Figure \ref{fig_exp_train_curves} plots training curves (only test accuracy is shown) for varying numbers of layers on MNIST, Fashion MNIST and CIFAR10, averaged over 10 runs. We compare a CNN with i.i.d.~prior to a CNN with a structured weight prior that combines both the feature specific prior (`Feats') for final layer weights, and probabilistic Gabor prior for first layer filters. We also include a fc NN with i.i.d.~prior.

For most plots, fc NN~i.i.d. $<$ CNN~i.i.d. $\leq$ CNN Gabor Feats, both in terms of convergence speed and best test accuracy. This is a significant result, providing evidence that better weight priors can benefit non-Bayesian NNs when used as initialising distributions.
Figure \ref{fig_exp_train_curves} also provides a way to visualise the relative benefit of architectural vs weight priors. In general the architectural prior (fc NN~i.i.d. vs CNN~i.i.d.) has a larger impact than the weight prior (CNN~i.i.d. vs CNN~Gabor~Feats).
%
%
%
%
Ablations for single-layer CNNs on MNIST are provided in appendix figure \ref{fig_exp_train_curves_ablation}, which show: 1) Performance is worse when $\sigma_g>0$. 2) Both the Gabor prior and final-layer prior in isolation deliver improvements, though most of the benefit comes from the Gabor prior.



\section{Related Work}

Several works have considered how to design useful functional priors with Bayesian NNs \citep{Flam-Shepherd2017,sun2019,Coker2019,Pearce2019}. These have generally considered fully-connected NNs, encoding granular properties of a function such as length scale and amplitude, often taking inspiration from Gaussian processes. 
%
%
Deep image prior \citep{Ulyanov2020}, showed that using features produced by randomly (i.i.d.) initialised CNNs can give surprisingly good performance on certain inverse vision tasks. This offers strong evidence that the architectural priors embedded in CNNs play a large part in their success. 

Several works have combined CNNs with Gabor filters. Alekseev \& Bobe (\citeyear{Alekseev2019}) proposed a CNN with Gabor functions integrated into the first-layer filters, such that the parameters controlling the output of the Gabor functions were learnt via gradient descent. Luan et al. (\citeyear{Luan2018}) proposed applying Gabor filters to process learnt filters. Ozbulak \& Ekenel (\citeyear{Ozbulak2018}) offered brief analysis initialising first-layer filters with Gabor functions, proposed as an alternative to transfer learning. By contrast, our work is the first to directly motivate and evaluate the Gabor function as a general-purpose prior, designed from a probabilistic standpoint.


\section{Discussion}
\label{sec_discuss}

Selecting a good architectural prior is an important step in deep learning, whilst weight priors are usually left vague --- often independent Gaussian distributions. This paper has asked the question: \textbf{can we do better than these vague weight priors?}
%
%
Whilst this question is of relevance across deep learning, the field of Bayesian deep learning, which requires a formal definition of the prior distribution within its models, is excellently placed to tackle it.

Indeed, there is ongoing debate on this topic. Following a comprehensive evaluation of Bayesian NNs (where they showed underwhelming performance), Wenzel et al.~(\citeyear{Wenzel2020}) concluded that further work on weight priors was a priority. Conversely, Wilson \& Izmailov (\citeyear{Wilson2020}) argued that architectural priors are themselves sufficient, and it may be superfluous to structure weight priors within these.
This paper shows that even a conceptually simple weight prior inspired by well-known image-processing techniques can lead to meaningful improvements in the functional prior. 

It may be that these prior improvements do not translate to better Bayesian \emph{posterior predictive} distributions, though section \ref{sec_exp_traincurve} provided some evidence that they are helpful in standard non-Bayesian learning.
%
Alternatively, the prior may yield significant improvements for the exact Bayesian posterior, but these may not translate into practical benefits if the approximate inference algorithm used is insufficiently accurate.
It may also be the case that in settings with larger datasets, there is sufficient data to completely overwhelm the effect of the weight prior.
We leave investigation of these issues to future work. In the meantime, our results have highlighted that thinking more deeply about weight priors is a promising direction for further exploration.


%



%
%

\bibliography{images/library}
\bibliographystyle{icml2020}

\newpage
\appendix
\section{Appendix}

\begin{algorithm}[h]
   \caption{Applying the Gabor prior to RGB Colour Channels}
   \label{alg_rgb_gabor}
\begin{algorithmic}
   \STATE {\bfseries Input:} A draw of Gabor parameters $\theta_g ,\sigma , \lambda,\psi, \gamma $, an RGB $\text{Filter} \in \mathbb{R}^{f_w \times f_w \times 3}$.
   \vspace{0.1in}
   \STATE $\text{P}_{bw} \sim \mathcal{U}(0,1)$
   \FOR{$i=1$ {\bfseries to} $3$}
   \IF{ $\text{P}_{bw}>0.3$} 
   \STATE $\beta \sim \mathcal{U}(-1,1)$ \# colour filter
   \ELSE
   \STATE $\beta \sim \mathcal{U}(0.8,1)$ \# b \& w filter
   \ENDIF
   \STATE $\text{Filter}[:,:,i] = \beta \times g(:,:)$  \# Gabor function eq. \ref{eq_gabor}
   \ENDFOR
\end{algorithmic}
\end{algorithm}

\begin{algorithm}[h]
   \caption{Feature Specific Priors}
   \label{alg_feats}
\begin{algorithmic}
   \STATE {\bfseries Input:} 20 samples of each class, $X_{i,j}$, for $i \in [0,1, \hdots, 20]$ and $j\in[0,1, \hdots, n_\text{classes}]$, a sample of $\theta_{\lnot \text{final}}$, and a NN with $h$ final-layer hidden nodes. Let $\text{NN}_k(X_{:,j},\theta_{\lnot \text{final}})$ be the $k^\text{th}$ hidden feature in the final layer for all examples of class $j$. Also define a vector to store the mean activation of each class, $\mu \in \mathbb{R}^{n_\text{classes}}$.
   
   \vspace{0.1in}
   
   \FOR{$k=1$ {\bfseries to} $h$}
   \FOR{$j=1$ {\bfseries to} $n_\text{classes}$}
   \STATE \# mean activation at hidden node $k$ for class $j$ 
   \STATE $\mu_{j} = \text{mean}(\text{NN}_k(X_{:,j},\theta_{\lnot \text{final}}))$ 
   \ENDFOR
   \STATE \# normalise to get \textit{relative} activations
   \STATE $\mu_{} = \mu_{} - \text{mean}(\mu_{})$ 
   \STATE \# weight prior for connections from hidden node $k$
   \STATE $\theta_{\text{final},i} \sim \mathcal{N}(\mu_{},0.1 I)$
   \ENDFOR
\end{algorithmic}
\end{algorithm}

\begin{figure}[h!]
\vskip 0.1in
\begin{center}
\centerline{}
\includegraphics[width=0.32\columnwidth]{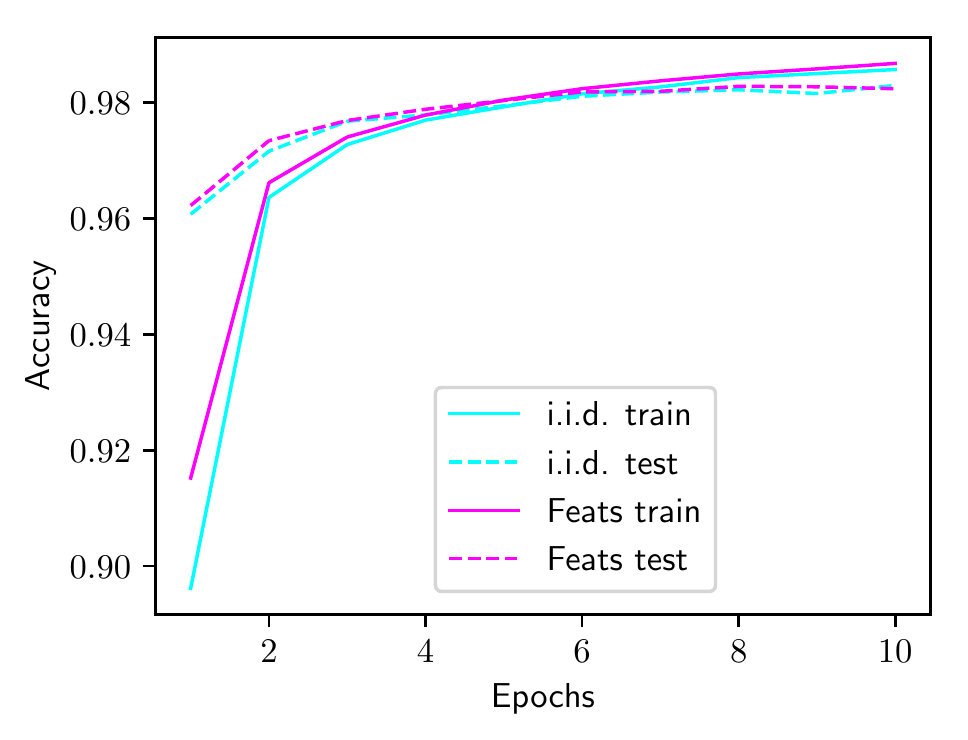}
\includegraphics[width=0.32\columnwidth]{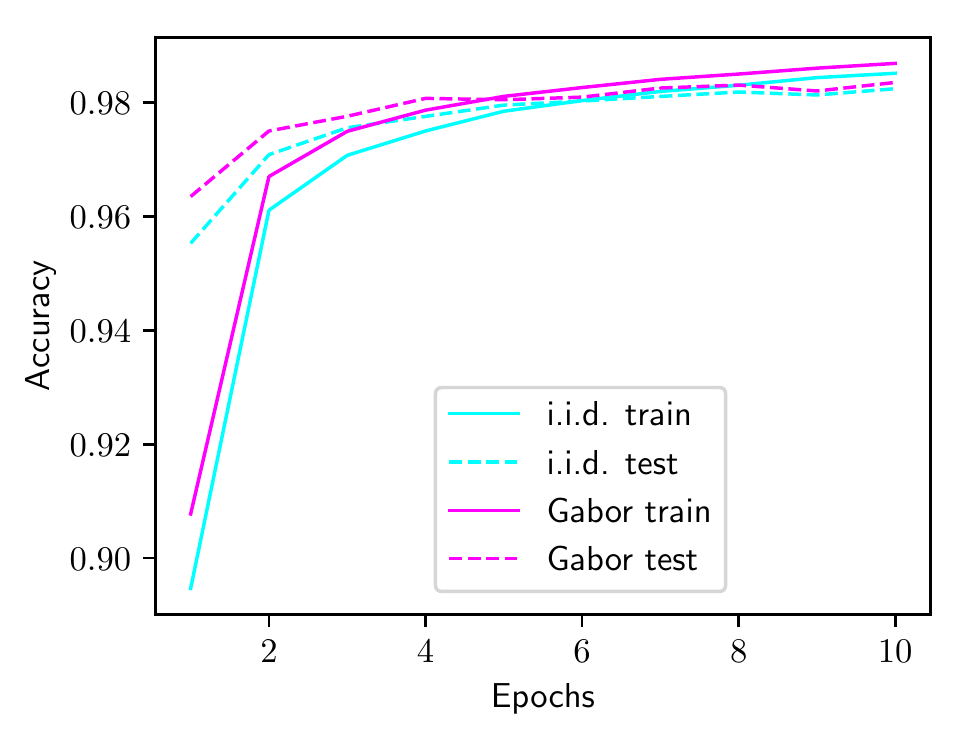}
\includegraphics[width=0.32\columnwidth]{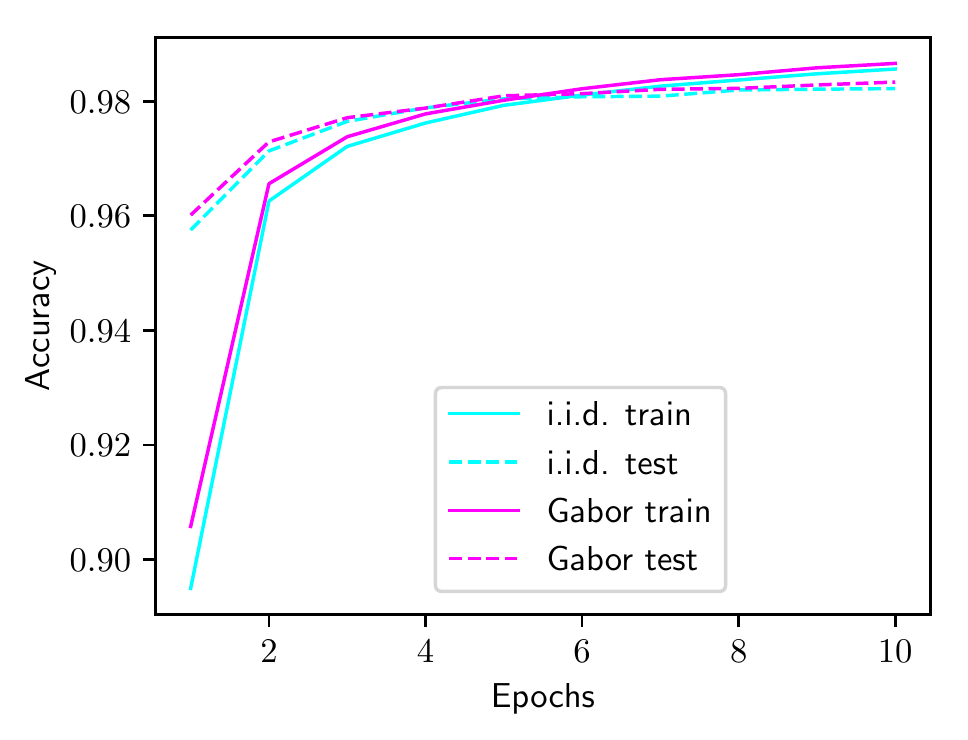}
\put(-65,62){\small Gabor, $\sigma_g=0.02$ }
\put(-142,62){\small Gabor, $\sigma_g=0.0$}
\put(-213,62){\small Features only}

\caption{Mean of five training runs for MNIST on a single-layer CNN, i.i.d.~vs structured priors where only one element of the structuring is used (either only features, or only Gabor filters).}
\label{fig_exp_train_curves_ablation}
\end{center}
\vskip -0.2in
\end{figure}

\end{document}